# Fair Patient Model: Mitigating Bias in the Patient Representation Learned from the Electronic Health Records

Sonish Sivarajkumar[1], Yufei Huang[1,3], Yanshan Wang[1,4,5,6] *

[1]Intelligent Systems Program, School of Computing and Information, University of Pittsburgh, Pittsburgh, PA; [2]Department of Medicine, University of Pittsburgh, Pittsburgh, PA; [3]Department of Electrical and Computer Engineering, University of Pittsburgh, Pittsburgh, PA; [4]Department of Health Information Management, University of Pittsburgh, Pittsburgh, PA; [5]Department of Biomedical Informatics, University of Pittsburgh, Pittsburgh, PA; [6]Clinical and Translational Science Institute, University of Pittsburgh, Pittsburgh, PA;

## Abstract

Objective: To pre-train fair and unbiased patient representations from Electronic Health Records (EHRs) using a novel weighted loss function that reduces bias and improves fairness in deep representation learning models.

Methods: We defined a new loss function, called weighted loss function, in the deep representation learning model to balance the importance of different groups of patients and features. We applied the proposed model, called Fair Patient Model (FPM), to a sample of 34,739 patients from the MIMIC-III dataset and learned patient representations for four clinical outcome prediction tasks.

Results: FPM outperformed the baseline models in terms of three fairness metrics: demographic parity, equality of opportunity difference, and equalized odds ratio. FPM also achieved comparable predictive performance with the baselines, with an average accuracy of 0.7912. Feature analysis revealed that FPM captured more information from clinical features than the baselines.

Conclusion: FPM is a novel method to pre-train fair and unbiased patient representations from EHR data using a weighted loss function. The learned representations can be used for various downstream tasks in healthcare and can be extended to other domains where bias and fairness are important.

---

* Corresponding author: yanshan.wang@pitt.edu

# 1. Introduction

Electronic Health Records (EHRs) have revolutionized the way patient data is collected and stored in healthcare settings. EHRs contain a wealth of patient data, such as patient demographics, medical history, laboratory results, clinical notes, and other information, which provides healthcare professionals with more comprehensive information about patients. The vast amounts of patient information stored in EHRs can enable clinicians to make more informed decisions about patient care (1). EHRs have also enabled data-driven predictions of drug effects and interactions, patient diagnosis, and other applications (2), paving the way for precision medicine and enabling tailored treatments based on individual patient characteristics, leading to improved patient outcomes.

Machine learning models are capable of learning patterns and relationships in EHR data to assist clinicians and healthcare professionals in making more informed decisions about patient care. Clinical machine learning often relies on feature selection and data representation techniques to identify the most relevant features for each clinical application. However, manual feature selection can be time-consuming and labor-intensive, making it difficult to identify the relevant features for each clinical application, particularly given the large size of EHR datasets(3). Additionally, this approach may reduce the generalizability of the framework, as the selected features can only be applied to that particular dataset or application, limiting the potential for transferability and scalability(4). Another drawback of manual feature selection is that it may lead to the exclusion of important clinical features that could have high predictive power. Consequently, specialized algorithms have been proposed to effectively represent EHR data, making them more effective in analyzing and predicting clinical outcomes across different patient populations and applications (5, 6).

Representation learning is a type of machine learning that learns the fundamental correlations among the data points and represents them in a lower-dimensional space(7). Deep patient representation learning utilizes deep learning algorithms to automatically extract and learn features from large and complex EHR data. The resulting representation effectively encapsulates significant clinical information, comprising medical history, vital signs, medication records, and laboratory findings. This representation can be used for various downstream healthcare applications, such as patient stratification(5), disease diagnosis (8), and drug discovery (9). Numerous deep patient representation learning models have been proposed in the literature that utilizes machine learning algorithms like autoencoders(10), Recurrent Neural Networks (RNN)(11), and graph neural networks(12). While these models have the potential to offer significant benefits in improving patient outcomes, they may suffer from biases that can lead to negative healthcare outcomes(13). For instance, if the training data mostly comprises patients from a particular demographic group, such as Caucasians, the resulting representations may not generalize well to patients from other groups like African Americans or Asians. Additionally, if the training data does not represent the actual population distribution, the model may learn spurious correlations that do not reflect the underlying causal relationships between clinical features and outcomes(14). Thus, it is essential to investigate the bias and fairness issues in deep patient representations and propose new models that can create unbiased patient representations.

In this paper, we investigate the presence of gender bias in deep patient representation learning models, and propose a novel unbiased patient representation model that utilizes an autoencoder architecture. The proposed "Fair Patient Model(FPM)" aims to generate unbiased and generalized patient representations that can be utilized for multiple downstream applications like patient mortality prediction, patient stratification, etc. FPM employs a customized loss function called the weighted reconstruction loss, which calculates the mean squared error loss for each patient and weights it by the inverse of its class frequency. We evaluate the model using the MIMIC-III dataset to predict patient mortality across different subpopulations, and demonstrate that the FPM representations outperform both deep patient representation models and common debiasing methods in terms of fairness scores.

## 2. Related Works

In recent years, there has been extensive research on deep learning models for generating patient representations. One of the most popular models is the Stacked Denoising Autoencoder (SDAE) model, which is capable of learning the complex relationships among the patient features and creating compact and informative patient representations(15). It uses an encoder-decoder stack to learn a set of weights that can project high-dimensional patient data onto a low-dimensional space while preserving the most important information. Deep Patient is one of the first works in deep patient representation learning model that uses SDAE model to learn low-dimensional representations of patients(16). Another study conducted by Che et al. used the SDAE model to learn patient representations from EHR data and demonstrated that these representations can accurately predict patient outcomes and identify subgroups of patients with similar clinical characteristics. However, these models can suffer from biases that can lead to disparities in access to healthcare(17).

Bias can occur in various forms, such as gender and ethnic bias, which can result in inaccurate predictions and treatment recommendations(18, 19). The term bias refers to systematic errors in data that can lead to incorrect predictions and decisions. Bias can be caused by factors like imbalanced training data, data quality issues, and inherent biases in healthcare systems, leading to inaccurate and unfair predictions, particularly for underrepresented populations(13, 20). Bias can stem from multiple sources, such as data collection and curation practices, historical and societal factors, and model design and training algorithms. Fairness, on the other hand, refers to treating individuals equitably regardless of their demographic characteristics. Studies have shown that gender and racial bias exist in healthcare, with female patients and patients from racial and ethnic minority groups experiencing a lower quality of care and worse health outcomes compared to their male and non-minority counterparts. For example, a study conducted by Hoffman et al. found that African American patients were less likely to receive appropriate treatment for their pain compared to white patients(19).

To address these biases, several techniques have been proposed in the literature. One approach is to use data augmentation techniques to balance the representation of different groups in the training data. For instance, a study by Suresh et al. (21) uses data augmentation techniques like oversampling to balance the representation of different gender and racial groups in the training data, which leads to fairer patient representations. Another approach is to use regularization techniques to reduce the impact of demographic characteristics on the patient representations.

For example, the method proposed by Sweeney et al. (15) uses demographic-aware regularization to reduce bias by accounting for the demographic distribution of the training data. Reweighting sampling is another commonly used preprocessing technique for addressing bias in datasets, particularly when working with imbalanced data(22). This technique involves adjusting the weights of individual samples in a dataset to account for differences in their representation or importance. This approach helps to address the issue of bias by ensuring that the model is exposed to a more balanced set of examples. Reweighting is a popular bias mitigation technique that involves adjusting the weight of each sample in the dataset based on its demographic attributes to reduce any imbalances in the data(23). This technique aims to address biases in the dataset by giving more weight to underrepresented samples and less weight to overrepresented samples, which can help to reduce the influence of bias on the predictive model.

Despite the growing interest in clinical AI fairness, there is still limited research that combines deep patient representation with fairness. Past studies in this field have primarily focused on mitigating bias in the dataset or downstream predictions, rather than tackling the fundamental issue of bias in patient representation learning. A study by Liu et. al employs a transformer-based attention mechanism that addresses bias in structured EHR data(24). Our framework combines both structured EHR and clinical notes and generated generalized patient representations themselves, ensuring that downstream models built on these representations are also unbiased. This is one of the first research works studies in the foundational AI models in the clinical domain that prioritizes unbiased representations and fairness across different subpopulations.

# 3. Methods

## 3.1. Dataset

In this study, we used the MIMIC-III (Medical Information Mart for Intensive Care III) dataset, which is a large, de-identified database of electronic health records (EHRs) collected from patients admitted to the Beth Israel Deaconess Medical Center between 2001 and 2012(25). The dataset contains clinical data from over 40,000 patients, including structured data such as chart events, CPT events, procedures, lab events, and prescription information, as well as demographic information such as ethnicity, gender, marital status, and insurance, and patient age at the time of admission. The MIMIC-III dataset is publicly available under the data use agreement.

The dataset covers a wide range of clinical conditions and procedures, with a total of 14,199 distinct ICD-9 diagnosis codes and 3,871 distinct ICD-9 procedure codes. The dataset also contains 2,083,180 clinical notes written by various healthcare providers during the patients' stays. We used a subset of the MIMIC-III dataset that consists of 35,351 adult patients (age ≥ 18 years) who had at least one note during their first admission. The average length of stay for these patients was 10.3 days (s.d. = 12.4) and the in-hospital mortality rate was 11.5%.

## 3.2. EHR Processing

*Structured Data*

The structured EHR data underwent several preprocessing steps to ensure that the data was clean and suitable for downstream analysis. Firstly, irrelevant columns (the patient ID, Admission ID, etc.) and duplicate rows were removed to reduce unnecessary noise and redundancy. Then, the continuous variable columns were normalized to achieve a mean of 0 and a standard deviation of 1. Finally, categorical variables were transformed into numerical values using one-hot encoding, which is a common technique for dealing with categorical data in machine learning.

We used structured data from the last encounter for each patient. Specifically, we extracted the demographics, clinical features, and descriptors from the following files: Admissions, Patients, Chart Events, CPT Events, Procedures, Lab Events, and Diagnosis. To ensure that we had high-quality and relevant data for analysis, we filtered out features that had less than 70% of the values. This step was taken to reduce the potential impact of missing values on the analysis and to ensure that the features we used were representative of the patient's clinical information. Further, to address missing values within the structured EHR data, imputation was conducted by replacing them with the median value for each respective column. This approach aimed to ensure that the models could still learn from the available data while minimizing the impact of missing values on the overall analysis. After preprocessing, the structured EHR data consisted of 16,865 features.

*Clinical Notes*
In order to integrate unstructured data into our patient representations, we utilized Latent Dirichlet Allocation (LDA) for topic modeling to extract latent topics from clinical notes. To prepare the clinical notes data for topic modeling, the first step involved removing negations from the text. Negations can be challenging for topic modeling algorithms as they can alter the intended meaning of the text and produce inaccurate results. For example, the statement "no history of diabetes" has a different context and meaning than "history of diabetes". We employed the NegEx algorithm(22) to identify and remove negations from the text . Secondly, we employed stemming to convert the clinical notes into their base forms. Stemming is a technique that reduces the derived words to their base or root form. This reduces the size of the vocabulary and enables the algorithm to better recognize and group related terms.

*Latent Dirichlet Allocation (LDA)*
Topic modeling is a machine learning technique that automatically analyzes text data to determine clusters of words that best represent the topics or themes of a set of documents (26). LDA is a generative model that assumes each document is a mixture of topics and each topic is a distribution of words(27). This technique has demonstrated success in capturing the underlying concepts in clinical notes for downstream tasks(28).

LDA can capture the underlying concepts in clinical notes and help integrate unstructured data into patient representations. We used the gensim(23) library to implement LDA on clinical notes. To optimize the performance of the LDA algorithm, we used a grid search approach to find the best hyperparameters, such as the number of topics and the Dirichlet priors. Our final LDA model used 100 topic probabilities, with a perplexity score of 11.0058. The perplexity score measures how well the model fits the data, with lower values indicating better fit4. After identifying the topics in the clinical notes, we used count vectorization to represent each patient's notes as a vector of topic weights. This approach involves counting the number of occurrences of each topic in the patient's notes and dividing by the total number of words in the notes.

*Integration*

Following the LDA topic modelling, we combined the derived topic weights with the structured data to generate the final patient representation. Patients who lacked clinical notes were excluded from the dataset. It should be noted that it is not uncommon for multiple encounters to take place over consecutive days or admissions, leading to the aggregation of these encounters into a single patient representation(29). In this study, such aggregation was achieved by computing the mean value for each feature across all encounters within a patient's record. The resulting patient representation was transformed into a table containing 35,351 patients and 16,965 columns, where each row corresponded to a distinct patient in the dataset and each column denoted a specific feature.

## 3.3. Distribution of Gender

One of the important tasks in clinical decision support systems is to predict the mortality risk of patients within a certain time span, such as 30-day, 60-day, 90-day, or 1-year mortality. This can help clinicians to identify high-risk patients and provide appropriate interventions. However, the accuracy and fairness of such predictions depend on the quality and representativeness of the patient data used to train the predictive models. We evaluate deep patient representations to predict the mortality within each time span and examine whether the gender bias in the dataset is reflected in the outcome prediction. In this section, we investigate the presence of gender bias in the dataset that we use for our experiments, which is a sample of 35,351 patients obtained after preprocessing the MIMIC-III dataset.

We analyzed the distribution of female and male patients in the dataset with respect to the mortality target variable for each time span. The overall distribution of females and males in our dataset was found to be 44% and 56%, respectively. However, since an equal distribution of gender in both categories (deceased or alive patients) cannot be assumed, we analyzed the gender distribution of patients in each mortality category. Figure 1 shows the number of patients who were either alive (0) or deceased (1) in each mortality category, stratified by gender.

For the 30-day mortality category, the total number of female patients who were alive was 12,798, while the number of deceased female patients was 2,552. The corresponding numbers for male patients were 16,913 and 3,088, respectively. The 60-day mortality category showed a similar pattern, with a total of 12,387 female patients and 16,404 male patients alive, while 2,963 female patients and 3,597 male patients were deceased. In the 90-day mortality category, 12,108 female patients and 16,020 male patients were alive, while 3,242 female patients and 3,981 male patients were deceased. Finally, the 1-year mortality category showed that 10,765 female patients and 14,388 male patients were alive, while 4,585 female patients and 5,613 male patients were deceased. This indicates that female patients have a higher mortality rate than male patients within the same time span.

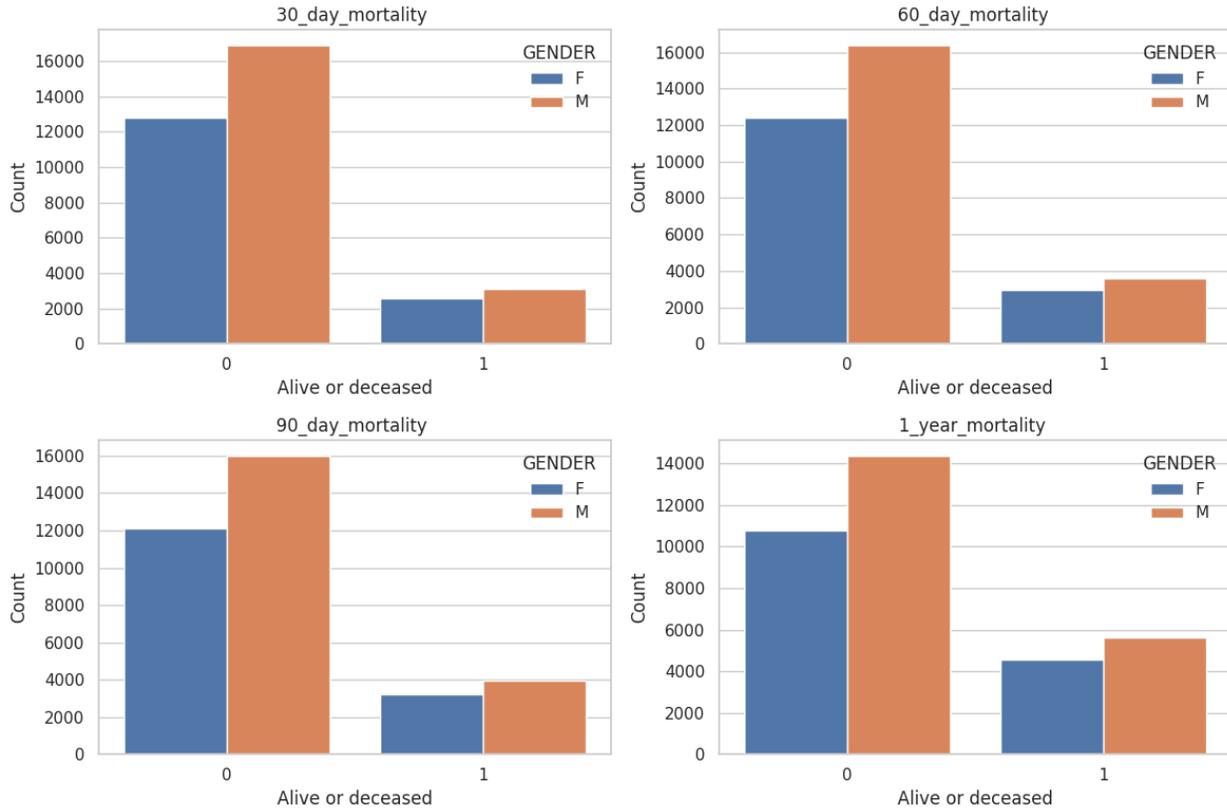

Figure 1: Imbalance in data across gender and mortality labels. Here, 0 means alive and 1 means deceased.

This reveals that there is a noticeable data bias in the mortality outcome across gender within the initial dataset. For each mortality category, the percentage of deceased female patients is higher than the percentage of deceased male patients, implying that female patients have a higher mortality rate than male patients within the same time span. Therefore, it is crucial to examine whether this bias affects the patient representations and the downstream outcome prediction tasks.

## 3.4. Patient Representations

In the baseline patient representation model, we use a three-layered Stacked Denoising AutoEncoder (SDAE) architecture, as used in Deep Patient (16), to learn low-dimensional, dense vectors representing EHR subsequences(16). An SDAE is a neural network model that learns a compressed representation of high-dimensional input data by removing noise and reconstructing the original data. It is comprised of several layers of denoising autoencoders, wherein each layer learns a compressed representation of the input data by encoding it into a lower-dimensional space and then decoding it back into the original space. Denoising serves as a regularization technique in the model to prevent overfitting and enhance the robustness of the learned features. The denoising autoencoder approach adds noise to the input data and trains the model to reconstruct the original, noise-free input. The noise is added by masking a portion of the input features, effectively setting them to zero, and then training the model to reconstruct the original output.

$$\tilde{x} = x \odot m$$

where x is the original input data, $\tilde{x}$ is the noisy input data, m is a binary mask vector with probability p of being 1, and $\odot$ is the element-wise product.

The training of the SDAE is performed in an unsupervised manner, where the output of the previous layer is fed as input to the next layer, thereby creating a hierarchical representation of the input data. A single layer of a denoising autoencoder can be represented as:

*Encoding:*

$$h = f(\tilde{x} \cdot W + b)$$

where $\tilde{x}$ is the input noisy data, W is the weight matrix, b is the bias vector, h is the hidden representation, and f is the activation function.

*Decoding:*

$$x' = g(h \cdot W' + b')$$

where x' is the reconstructed input data, W' is the weight matrix for the decoding layer, b' is the bias vector for the decoding layer, h is the hidden representation from the encoding layer, and g is the activation function for the decoding layer.

The unsupervised training of the three-layered SDAE was performed using backpropagation with stochastic gradient descent. The gradients with respect to the parameters in each layer are computed to update the network weights. Upon training, the compressed representation of the input EHR data was obtained by passing it through the encoding layers of the SDAE. The SDAE architecture utilized in this study was designed with a hidden layer and final layer size of 500, consisting of 500 neurons in each layer. Prior research has shown that using the same parameters for all layers, including the hidden and output layers, results in a similar performance to using distinct parameters for each layer, while also simplifying the evaluation process(22). The SDAE model generated patient representations as a vector of 500 features. More details about the model, including the architecture and the hyperparameters, are provided in the appendix.

## 3.5. Downstream tasks

To evaluate the effectiveness and fairness of the patient representations generated by the base patient representation learning model(SDAE), we performed four downstream tasks for predicting mortality within different time spans: 30 days, 60 days, 90 days, and 1 year. However, as we discussed in Section 3.3, there is a noticeable data bias in the mortality outcome across gender within the initial dataset, which implies that female patients have a higher mortality rate than male patients within the same time span. Therefore, it is crucial to examine whether this bias affects the downstream outcome prediction tasks. For each task, we utilized the patient representations as inputs to various classifiers such as random forest, decision tree, and XGBoost.

The distribution of deceased and alive patients for each task is illustrated in Figure 2. We can observe that the number of deceased patients is much higher than the number of alive patients in all tasks, which indicates an imbalance in the dataset. This imbalance can affect the performance and fairness of the classifiers, as they may tend to predict more patients as deceased than alive.

We can also observe that the number of deceased patients decreases as the time span increases, while the number of alive patients increases. This means that the mortality prediction becomes more challenging for longer time spans, as there are fewer positive examples and more negative examples. The classifiers may have less information to learn from the deceased patients and more noise to filter out from the alive patients. Therefore, we need to use more robust and accurate tree-based classifiers, which have been proven to perform better in highly imbalanced data(30) .

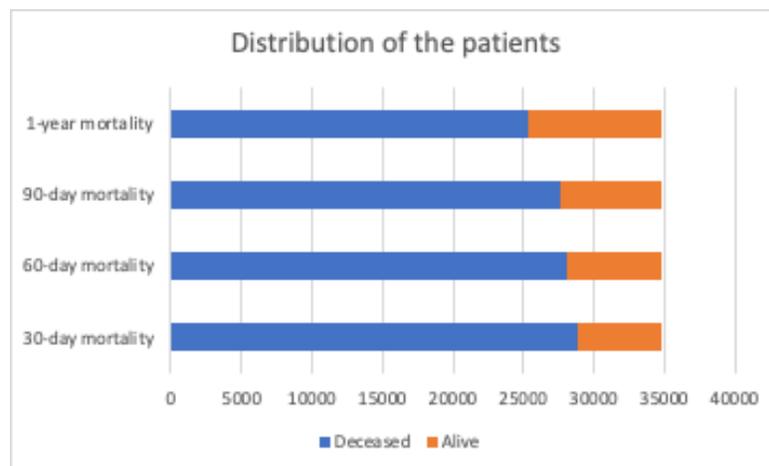

Figure 2: Distribution of patients in each of the 4 tasks

## 3.6 Fairness Evaluation

In addition to assessing the classifier's accuracy in predicting mortality, we conducted a fairness evaluation to examine potential biases in the patient representations used for prediction. Our evaluation focused on gender and involved analyzing the classifier's performance across different gender groups. Initially, we determined the proportion of positive outcomes for each group, representing the percentage of individuals within each group who received a positive outcome, such as being predicted as deceased. Subsequently, we employed three fairness metrics to evaluate the classifier's fairness: demographic parity ratio(31), equality of opportunity difference(32), and equalized odds ratio(33).

The demographic parity ratio quantifies the disparity in overall error rates among various subgroups. It is computed by comparing the percentage of positive outcomes for the privileged group (e.g., non-minorities) to that of the unprivileged group (e.g., minorities). Specifically, for gender, the demographic parity ratio is calculated as:

$$Demographic\ parity\ ratio = \frac{P(Y = 1|G = male)}{P(Y = 1|G = female)}$$

In the equation, Y represents the outcome variable, and G represents the gender variable. A demographic parity ratio of 1 suggests no difference in overall error rates across subgroups. A value greater than 1 indicates that females have a higher error rate than males, while a value less than 1 suggests that males have a higher error rate than females.

The Equality of Opportunity Difference(EOD) measures the discrepancy in false negative rates among different subgroups. It is determined by the difference between the percentage of positive outcomes for the unprivileged group (e.g., minorities) and the percentage of positive outcomes for the privileged group (e.g., non-minorities) among those who actually have a positive outcome. For gender, the Equality of Opportunity Difference can be calculated as:

$$EOD = P(Y = 0|G = female, Y' = 1) - P(Y = 0|G = male, Y' = 1)$$

Here, Y represents the outcome variable, Y' represents the true outcome variable, and G represents the gender variable. An EOD Difference of 0 signifies no disparity in false negative rates across different subgroups. A negative value suggests that males have a higher false negative rate than females, while a positive value suggests that females have a higher false negative rate than males.

The equalized odds ratio assesses the difference in both false positive and false negative rates across various subgroups. It is calculated by comparing the percentage of positive outcomes for the privileged group to that of the unprivileged group among those who have the same true outcome. Specifically, for gender, the equalized odds ratio can be expressed as:

$$Equalized\ odds\ ratio = \frac{P(Y = 1|G = male, Y' = 0)}{P(Y = 1|G = female, Y' = 0)} X \frac{P(Y = 0|G = male, Y' = 1)}{P(Y = 0|G = female, Y' = 1)}$$

In the equation, Y represents the outcome variable, Y' represents the true outcome variable, and G represents the gender variable. An equalized odds ratio of 1 indicates no disparity in both false positive and false negative rates across different subgroups. A value greater than 1 suggests that females have a higher error rate than males in both cases, while a value less than 1 suggests that males have a higher error rate than females in both cases.

## 3.6. Fair Patient Model(FPM)

We introduce a novel approach to construct unbiased and fair patient representation through the incorporation of a re-weighting loss function in the SDAE model. Our aim is to enhance the representation of underrepresented groups in the patient data, thereby improving the fairness performance of subsequent tasks, including but not limited to mortality prediction.

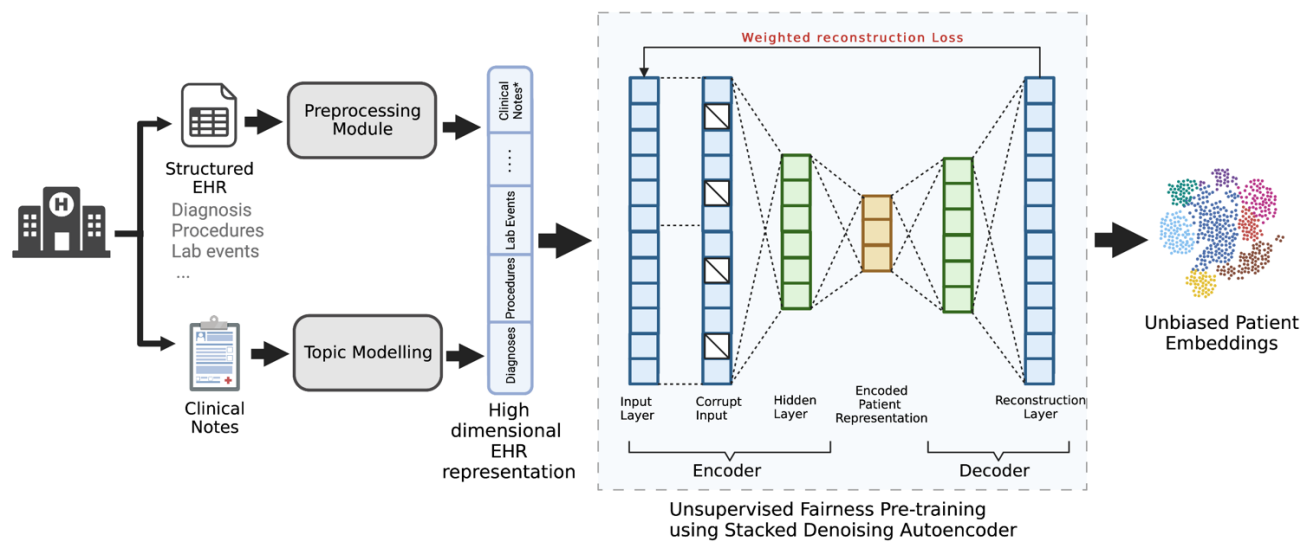

Figure 3: An overview of the patient representation pipeline. The dashed box shows FPM with weighted reconstruction loss.

### 3.6.1. FPM architecture

The proposed FPM method involves incorporating a re-weighting loss function and a new custom loss function, named "weighted reconstruction loss", into the base SDAE model. The aim of this method is to enhance model performance by improving the representation of underrepresented classes in the patient data. To implement the weighted loss function, we first compute the weight of each demographic class by taking the inverse of their frequency. These weights are then used to modify the loss function, so that the contribution of each example to the loss is weighted by the inverse of its class frequency. Finally, we calculate the weighted reconstruction loss as the sum of the products of each example's weight and its corresponding reconstruction loss.

The reconstruction loss, which is commonly used in autoencoder models, is defined as the mean squared error (MSE) between the original input data and its reconstruction. It is calculated as the Euclidean distance between the original input data (x) and its reconstruction (x') obtained from the latent representation (z) through the decoder.

Therefore, the reconstruction loss is defined as:

$$\text{Reconstruction loss} = \frac{||\text{x} - \text{x}'||^2}{n_{samples}}$$

where $x'$ is the reconstruction of x obtained from z, $n_{samples}$ is the number of samples, and $||.||^2$ denotes the squared Euclidean distance.

The proposed weighted reconstruction loss function is then defined as:

$$\text{Weighted reconstruction loss} = \frac{1}{n_{samples}} \sum_{i=1}^{n_{samples}} w_i ||x_i - x_i'||^2$$

where $w_i$ is the weight of the i-th example, which is calculated as

$$w_i = \frac{1}{freq(c_i)},$$

where, $freq(c_i)$ is the frequency of the i-th example's class label $c_i$ in the dataset. The optimization algorithm used for training the re-weighted deep patient model is stochastic gradient descent (SGD).

The encoder and decoder networks both use sigmoid activation functions, defined as

$$\sigma(z) = \frac{1}{1+e^{-z}}, \text{ and}$$

Rectified Linear Unit (ReLU) activation functions in their linear layers, defined as

$$\text{ReLU}(z) = \max(0, z)$$

The sigmoid activation function was applied to the output layer of the encoder and decoder to yield values within the [0, 1] range, allowing for probability or scaled value interpretation. Meanwhile, ReLU activation function was used in the hidden layers of the encoder and decoder to introduce non-linearity and sparsity in the learned representations.

During training, the input data is corrupted with a noise factor to encourage the model to learn more robust representations, defined as

$$x_{noise} = x + noise_{factor} * N(0,1),$$

where x is the original input data, noise_factor is the corruption factor (set to 5% in this case), and N(0, 1) is a Gaussian noise distribution with mean 0 and standard deviation 1. The loss function is optimized using stochastic gradient descent (SGD) algorithm, defined as

$$w_i = w_i - \text{learning\_rate} \times \nabla w_i,$$

where $w_i$ is the weight of the i-th example, learning_rate is the step size of the optimization algorithm, and $\nabla w_i$ is the gradient of the loss function with respect to $w_i$. The gradient is calculated using backpropagation through the network.

Thus, FPM enables the model to learn from the entire dataset while appropriately weighing each example based on its class frequency, thereby improving the model's ability to capture and represent the underlying patterns and relationships in the data.

### 3.6.2 Baselines

We conducted a comparative analysis of FPM, evaluating two different approaches. The first approach involved using a machine learning classifier with a baseline SDAE representation as described in 3.4. The SDAE representation was obtained by training a stacked denoising autoencoder (SDAE) on the original clinical features without any fairness constraints. The SDAE was composed of three hidden layers with 500 units each, and used a Tanh activation function and a mean squared error loss function. The SDAE was trained for 100 epochs with a batch size of 128 and a learning rate of 0.001. The output of the last hidden layer of the SDAE was used as the baseline patient representation. Then, we applied a machine learning classifier on the baseline representation to predict mortality in four time spans: 30 day, 60-day, 90-day, and 1 year. We limited our experiments to only XGBoost classifier for our experiments because we have already demonstrated its superior performance.

In the second approach, reweighting was used as a step to mitigate the effects of bias in the dataset before generating patient representations and applying a machine learning classifier for predicting mortality. Reweighting is a preprocessing technique that assigns weights to each instance in the dataset based on its sensitive attribute value and label, such that the weighted dataset satisfies demographic parity. We used the reweighting algorithm proposed by Kamiran and Calders(22), which computes the weights as follows:

$$w_i = \frac{P(y_i)}{P(y_i \mid s_i)}$$

where $w_i$ is the weight of instance i, $y_i$ is its label, and $s_i$ is its sensitive attribute value. We used gender as the sensitive attribute to reweight the original dataset. Then, we trained a SDAE on the reweighted dataset to get the weighted patient representations. We kept the same hyperparameters as the base SDAE. Finally, we used the same machine learning classifiers as in the first approach to predict mortality from the weighted representations

# 4. Results

## 4.1 Downstream classifiers

In this section, we compare the performance of different downstream classifiers on the mortality prediction tasks. We use three classifiers: Random Forest, Decision Tree and XGBoost. We train each classifier on the patient representations obtained from the FPM model and evaluate them on the test set. Table 2 shows the accuracy of each classifier on each of the four tasks: 30-day, 60-day, 90-day and 1-year mortality prediction. The table illustrates that the XGBoost classifier outperformed the other classifiers on all tasks, achieving the highest accuracy for each task. The Random Forest classifier performed second best, followed by the Decision Tree classifier. We also experimented with two other classifiers: Naive Bayes and Logistic Regression. However, these classifiers gave poor performance on the downstream tasks using the patient representations.

Difference in accuracy can be observed between the XGBoost and the other classifiers for all tasks, indicating that the XGBoost classifier was more effective in predicting patient outcomes based on the patient representations. Hence, for further evaluations, we used the XGBoost classifier as the downstream classifier for comparing our model with the baselines.

Table 1: Comparison of accuracy for different downstream classifiers on the mortality prediction tasks using FPM. The table shows the accuracy of three classifiers: Random Forest, Decision Tree, and XGBoost on four outcome prediction tasks: 30-day mortality, 60-day mortality, 90-day mortality and 1-year mortality.

| Task | Model | Accuracy |
| --- | --- | --- |
| 30 day mortality | Random Forest | 0.8060 |
|  | Decision Tree | 0.7856 |
|  | XGBoost | 0.8382 |
| 60 day mortality | Random Forest Classifier | 0.7686 |
|  | Logistic Regression | 0.7572 |
|  | XGBoost | 0.8130 |
| 60 day mortality | Random Forest | 0.7557 |
|  | Decision Tree | 0.7374 |
|  | XGBoost | 0.7950 |
| 1 year mortality | Random Forest | 0.6431 |
|  | Decision Tree | 0.6321 |
|  | XGBoost | 0.7086 |

## 4.2 Gender bias evaluation

In this section, we evaluate the fairness and performance of different models on mortality prediction tasks. We focus on the gender bias in unsupervised patient representations, as gender is a sensitive attribute that may affect the outcomes and treatment decisions of patients. We compare the base SDAE model, the FPM model proposed in this study, and the SDAE model combined with reweighting as a preprocessing data bias mitigation strategy. We use three fairness metrics: Demographic Parity Ratio, Equality of Opportunity Difference and Equalized Odds Ratio. We also use accuracy and AUROC as performance metrics. Table 1 shows the results of the comparison for four mortality prediction tasks: 30-day, 60-day, 90-day and 1-year. We discuss the results and implications of each task in the following subsections.

**Table 2**

Comparison of fairness and performance metrics for different models on mortality prediction tasks. The table shows the Demographic Parity Ratio, Equality of Opportunity Difference, Equalized Odds Ratio, Accuracy and AUROC for SDAE, FPM and reweighting + SDAE on four outcome prediction tasks: 30-day mortality, 60-day mortality, 90-day mortality and 1-year mortality. A Demographic Parity Ratio value of 1 indicates that there is no difference in the overall error rates across different subgroups. An Equality of Opportunity Difference value of 0 indicates that there is no difference in the false negative rates across different subgroups. An Equalized Odds Ratio value of 1 indicates that there is no difference in the false positive and false negative rates across different subgroups.

| Task | Model | Demographic Parity Ratio | Equality of Opportunity Difference | Equalized Odds Ratio | Accuracy | AUROC |
|---|---|---|---|---|---|---|
| 30 day mortality | SDAE | 0.1702 | 0.0300 | 0.1372 | 0.8382 | 0.6510 |
| 30 day mortality | FPM | 1.1643 | 0.0423 | 1.2785 | 0.8381 | 0.6486 |
| 30 day mortality | Reweighting + SDAE | 3.0530 | 0.0870 | 3.8289 | 0.8357 | 0.6470 |
| 60 day mortality | SDAE | 0.5561 | 0.2462 | 0.1998 | 0.8130 | 0.618 |
| 60 day mortality | FPM | 1.0809 | 0.0090 | 1.0787 | 0.8128 | 0.6580 |
| 60 day mortality | Reweighting + SDAE | 1.1316 | 0.0452 | 1.4242 | 0.8080 | 0.6448 |
| 90 day mortality | SDAE | 0.6568 | 0.1471 | 0.5059 | 0.7950 | 0.6533 |
| 90 day mortality | FPM | 1.1449 | 0.0468 | 1.2700 | 0.7952 | 0.6679 |
| 90 day mortality | Reweighting + SDAE | 1.9479 | 0.0853 | 0.0853 | 0.7909 | 0.6112 |
| 1 year mortality | SDAE | 1.0204 | 0.1027 | 0.7329 | 0.7086 | 0.6769 |
| 1 year mortality | FPM | 0.9657 | 0.0340 | 0.8290 | 0.7189 | 0.6898 |
| 1 year mortality | Reweighting + SDAE | 1.4948 | 0.0149 | 1.5852 | 0.6947 | 0.6086 |

For the 30-day mortality prediction task, the proposed model, FPM, achieved a higher Demographic Parity Ratio of 1.1643 and a higher Equalized Odds Ratio of 1.2785 than SDAE, indicating that the outcomes are more balanced across different subgroups. The Equality of Opportunity Difference of FPM is also higher than SDAE, suggesting that the model has better fairness in terms of false negative rates. The Accuracy and AUROC of FPM are slightly lower than SDAE, but the difference is negligible. Reweighting + SDAE obtained the highest Demographic Parity Ratio and Equalized Odds Ratio, but also the highest Equality of Opportunity Difference, indicating that the model is more biased in terms of false positive rates. The Accuracy and AUROC of reweighting + SDAE are also lower than FPM and SDAE.

For the 60-day mortality prediction task, the proposed model also outperformed SDAE in terms of Demographic Parity Ratio and Equalized Odds Ratio, showing that the model has better fairness in predicting mortality for different subgroups. The Equality of Opportunity Difference of FPM is almost zero, indicating that the model is equally fair in terms of false negative rates for both subgroups. The Accuracy and AUROC of FPM are almost identical to SDAE, indicating that the model does not sacrifice performance for fairness. Reweighting + SDAE achieved a slightly higher Demographic Parity Ratio than FPM , but a lower Equalized Odds Ratio and a higher Equality of Opportunity Difference than both FPM and SDAE, suggesting that the model is more biased in terms of false positive rates. The Accuracy and AUROC of reweighting + SDAE are also lower than FPM and SDAE.

For the 90-day mortality prediction task, the proposed model again achieved higher Demographic Parity Ratio and Equalized Odds Ratio than SDAE, demonstrating that the model is more robust and fair in predicting mortality for different subgroups. The Equality of Opportunity Difference of FPM is also higher than SDAE, showing that the model is fairer in terms of false negative rates. The Accuracy and AUROC of FPM are slightly higher than SDAE, suggesting that the model also improves performance. Reweighting + SDAE obtained a much higher Demographic Parity Ratio than both FPM and SDAE, but a much lower Equalized Odds Ratio and a much higher Equality of Opportunity Difference than both FPM and SDAE, indicating that the model is highly biased in terms of false positive rates. The Accuracy and AUROC of reweighting + SDAE are also significantly lower than FPM and SDAE.

For the 1-year mortality prediction task, the proposed model obtained a lower Demographic Parity Ratio of 0.9657 than SDAE, indicating that the outcomes are more proportional across different subgroups. The Equalized Odds Ratio and Equality of Opportunity Difference of FPM are also higher than SDAE, showing that the model is fairer in terms of both overall and conditional error rates. The Accuracy and AUROC of FPM are significantly higher than SDAE, indicating that the model also enhances performance. Reweighting + SDAE achieved a higher Demographic Parity Ratio than both FPM and SDAE, but a lower Equalized Odds Ratio and a lower Equality of Opportunity Difference than both FPM and SDAE, suggesting that the model is less fair in terms of both overall and conditional error rates. The Accuracy and AUROC of reweighting + SDAE are also significantly lower than FPM and SDAE.

Overall, the results suggest that the proposed model, FPM, outperformed SDAE in terms of fairness metrics for all mortality prediction tasks, while maintaining or improving accuracy and AUROC. These findings suggest that incorporating subpopulation-specific information during pre-training could result in more robust and fair patient model representations that capture important clinical features while mitigating potential biases. Reweighting + SDAE performed worse than both FPM and SDAE in terms of fairness metrics and performance metrics for all mortality prediction tasks, indicating that reweighting alone is not sufficient to address gender bias in unsupervised patient representations.

## 4.4 Feature Analysis

The table 3 shows the train loss, validation loss, and reconstruction loss for FPM and SDAE models. The overall train loss, validation loss, and reconstruction loss of your autoencoder are the average values of these metrics across all the features. They reflect the performance of your autoencoder as a whole, rather than for each individual feature. The train loss and validation loss are the mean squared errors between the predicted outcomes and the actual outcomes for the training and validation sets, respectively. For SDAE, the reconstruction loss is the mean squared error between the original feature values and the reconstructed feature values after passing through the encoder and decoder layers. For FPM, the reconstruction loss is the weighted mean square error based on the demographic distribution(gender).

We can see that the FPM achieved much lower train loss and validation loss than SDAE, indicating that FPM was able to fit the data better and generalize well to unseen data. This implies that FPM learned more robust and discriminative features that are useful for predicting patient outcomes. SDAE, on the other hand, had higher train loss and validation loss, suggesting that SDAE was overfitting the data and unable to capture the underlying patterns and relationships among the features.

The reconstruction loss for both models was very low, indicating that both models were able to preserve most of the feature information during the unsupervised pre-training phase. However, as we saw in Table 4, FPM had lower reconstruction errors for most of the features than SDAE, especially for the clinical topics. This means that FPM was able to reconstruct the features more accurately and faithfully than SDAE. SDAE may have compromised some feature information in order to reduce the prediction errors. Therefore, FPM had a better balance between feature preservation and outcome prediction than SDAE.

**Table 3**: Overall train loss, validation loss, and reconstruction loss of FPM and SDAE

| Clinical Feature | Train Loss | Validation Loss | Reconstruction Loss |
|---|---|---|---|
| FPM | $2.4762 \times 10^{-6}$ | $2.4765 \times 10^{-6}$ | $2.4761 \times 10^{-6}$ |
| SDAE | $4.3722 \times 10^{-4}$ | $4.3688 \times 10^{-4}$ | $2.4823 \times 10^{-6}$ |

To evaluate the quality of the learned features by FPM and SDAE, we calculated the reconstruction errors for each clinical feature during the unsupervised pre-training phase. The reconstruction error is defined as the mean squared difference between the original feature value and the reconstructed feature value after passing through the encoder and decoder layers. A lower reconstruction error indicates a better preservation of the feature information.

**Table 4**
    (a) The best feature reconstructions during unsupervised pre-training of FPM.

| Clinical Feature | FPM Reconstruction Error | Clinical Feature | SDAE Reconstruction Error |
|---|---|---|---|
| Renal Failure | $7.1453 \times 10^{-5}$ | Pleural Effusion | $3.6455 \times 10^{-5}$ |
| Head trauma-intracranial hemorrhage | $7.0019 \times 10^{-5}$ | Hyponatremia | $3.6252 \times 10^{-5}$ |
| Pneumonia | $6.912 \times 10^{-5}$ | Lung Lesion | $3.5970 \times 10^{-5}$ |
| Multiple Myeloma | $6.8464 \times 10^{-5}$ | Metastatic Lung Cancer | $3.5810 \times 10^{-5}$ |
| Topic 33: Electrolyte and Liver Tests | $6.7719 \times 10^{-5}$ | Popliteal Aneurysm Left | $3.5694 \times 10^{-5}$ |
| Topic 1: Abdominal Ultrasound Evaluation | $6.5829 \times 10^{-5}$ | Bone Marrow Transplant | $3.5665 \times 10^{-5}$ |
| Brain lesions | $6.5755 \times 10^{-5}$ | Rib Fracture | $3.556 \times 10^{-5}$ |
| Aortic Stanosis | $6.3381 \times 10^{-5}$ | Non-Small Cell Lung Cancer | $3.5154 \times 10^{-5}$ |
| Coronary Artery Disease | $6.3195 \times 10^{-5}$ | CPT_NUMBER_35661: Bypass Graft Procedures Other Than Vein | $3.2668 \times 10^{-5}$ |
| Spinal Sepsis | $6.2631 \times 10^{-5}$ | Vessel Disease | $3.1590 \times 10^{-5}$ |

(b) The worst feature reconstructions during unsupervised pre-training of FPM.

| Clinical Feature | FPM Reconstruction Error | Clinical Feature | SDAE Reconstruction Error |
|---|---|---|---|
| CPT_CD_99291.1: Critical care | 0.1830 | CPT_CD_99291.1: Critical care | 0.1825 |
| CPT_NUMBER_99232.0: Subsequent hospital care | 0.1418 | CPT_NUMBER_99232.0: Subsequent hospital care | 0.1736 |
| CPT_CD_99232.1: Subsequent hospital care | 0.1377 | CPT_CD_99232.1: Subsequent hospital care | 0.1353 |
| Topic 8: Valvular-Ventricular Ultrasound(Cardiac) 8 | 0.1350 | Topic 8: Valvular-Ventricular Ultrasound(Cardiac) | 0.1352 |
| CPT_CD_99223.1: Initial hospital care | 0.1309 | CPT_CD_99223.1: Initial hospital care | 0.1318 |
| CPT_CD_99222: Moderate inpatient/observation evaluation | 0.1139 | CPT_NUMBER_99223.0: Initial hospital care | 0.1137 |
| Topic 41: Sinus-Ischemia ECG | 0.1108 | CPT_CD_99222.1: Moderate inpatient/observation evaluation | 0.1132 |
| Topic 79: Blood Oxygen Level Monitoring | 0.1095 | Topic 41:Sinus-Ischemia ECG | 0.1118 |
| CPT_CD_99285: Emergency department visit | 0.1077 | Topic 63: Abdominal Pacemaker Placement | 0.1112 |
| CPT_NUMBER_36556.0: Insertion of non-tunneled centrally inserted central venous | 0.0969 | Topic 79: Blood Oxygen Level Monitoring | 0.1098 |

| catheter | | | |

Table 4 shows the best and worst feature reconstructions for both models. We can see that FPM achieved lower reconstruction errors than SDAE for most of the features, especially for the clinical topics derived from topic modeling. This suggests that FPM was able to capture the semantic relationships among the clinical features more effectively than SDAE. FPM also performed better than SDAE for some specific clinical features, such as renal failure, pleural effusion, head trauma-intracranial hemorrhage, and hyponatremia. These features are related to critical conditions that may affect patient outcomes and require careful attention. FPM was able to preserve their information with high fidelity, which may improve the predictive performance of the model.

On the other hand, SDAE performed slightly better than FPM for some CPT codes, such as 99291, 99232, and 99223. These codes are related to evaluation and management services that are commonly used in hospital settings. SDAE may have learned to reconstruct these features more accurately because they have higher frequency and variability in the data. However, these features may not be very informative or discriminative for predicting patient outcomes, as they are not specific to any disease or condition. Therefore, FPM may have prioritized other features that are more relevant and meaningful for the prediction task.

# 5. Conclusion and future work

This paper presented a novel method to pre-train fair and unbiased patient representations from EHR data using a weighted loss function. The proposed method, called Fair Patient - Model(FPM), aimed to address the bias and fairness issues that may arise in deep representation learning models and affect the quality and reliability of the learned features. We evaluated the performance of FPM on a sample of patients from the MIMIC-III dataset and compared it with conventional deep representation learning models, such as stacked denoising autoencoder combined reweighting bias mitigation strategy. We used four different prediction tasks to assess the usefulness and generalizability of the learned representations: 30-day mortality, 60-day mortality, 90-day mortality, 1-year mortality prediction. We also used three fairness metrics to measure the degree of bias and discrimination in the models: demographic parity, equality of opportunity difference, and equalized odds ratio.

The results showed that FPM achieved lower bias and higher fairness than the other models for all the prediction tasks. FPM also improved the predictive performance for 30-day mortality prediction, which is a critical and challenging task with a low proportion of positive instances. For the other tasks, FPM performed comparably with the other models, without compromising the feature quality or accuracy. Moreover, we conducted a feature analysis to examine the reconstruction errors and feature importance for each model. We found that FPM preserved more information for most of the clinical features, especially for the clinical topics derived from topic modeling. FPM also selected more relevant and meaningful clinical features for different prediction tasks, demonstrating its ability to capture the semantic relationships among the features.

This work contributes to the advancement of machine learning applications in healthcare by providing a novel method to pre-train fair and unbiased patient representations from EHR data. The learned representations can be used for various downstream tasks, such as patient outcome prediction, patient similarity analysis, patient cohort identification, and clinical decision support. The proposed method can also be extended to other domains where bias and fairness are important considerations.

*Limitation and Future work*
One limitation of our work is that we did not explicitly model the temporal dynamics of the patient's condition over time. We aggregated the features by computing the mean value across all encounters within a patient's record, which may lose important information about the trends, patterns, and variations of the clinical data. A possible direction for future work is to explore different ways of incorporating temporality into our model, such as using time-aware attention mechanisms or Temporal Convolutional Networks (TCNs). These methods could capture the sequential order and dependencies of the clinical events, as well as their recency and relevance to the prediction task. We expect that these methods could improve the performance and fairness of our model, as well as provide more interpretable and actionable insights for clinical decision making.

For future work, we plan to extend our framework to address other sources of bias and unfairness in EHRs, such as ethnicity, age, socioeconomic status, and comorbidities. We also plan to explore different ways to incorporate fairness constraints and objectives into the deep patient model, such as using multi-task learning, domain adaptation, or causal inference techniques. Moreover, we plan to conduct more comprehensive evaluations of the clinical utility and interpretability of the FPM representations using expert feedback and real-world applications.

Generative AI, which has gained increasing attention in recent years, offers a promising direction for future research on fair patient representation. By leveraging generative techniques, we could create more balanced and representative patient embeddings that capture the diversity and complexity of the patient population. Such capabilities could enable us to better identify and address the sources of bias and unfairness in our model, as well as to improve its generalizability and robustness across different settings and scenarios.

# Acknowledgments

The authors would like to acknowledge support from the University of Pittsburgh Momentum Funds, Clinical and Translational Science Institute Exploring Existing Data Resources Pilot Awards, the School of Health and Rehabilitation Sciences Dean's Research and Development Award, and the National Institutes of Health through Grant UL1TR001857.

# Appendix

**Fair Patient Model(FPM) architecture details**

In this section, we provide more details about the FPM model that we proposed in this paper. The FPM model is based on the stacked denoising autoencoder (SDAE) model, which is a popular unsupervised learning method for learning low-dimensional representations of high-dimensional data. The SDAE model and the FPM model have similar architectures, except that the FPM model uses a weighted reconstruction error loss function, while the SDAE model uses a mean squared error loss function. The weighted reconstruction error loss function allows the FPM model to balance the trade-off between accuracy and fairness, by assigning different weights to different input features according to their importance and sensitivity. The SDAE model does not consider the fairness aspect and treats all input features equally. We will explain the architecture and the hyperparameters of the FPM model in the following tables.

**Table 1.** The architecture of the FPM model. The model consists of a stacked denoising autoencoder with six linear layers and an input layer. The input and output layers have the same size as the number of features in the data. The encoder and decoder layers have the same size and activation function, except for the last decoder layer, which uses a Tanh activation function.

| Layer | Type | Size | Activation |
| --- | --- | --- | --- |
| Input | Linear | 100 | None |
| Encoder 1 | Linear | 500 | ReLU |
| Encoder 2 | Linear | 500 | ReLU |
| Encoder 3 | Linear | 500 | ReLU |
| Decoder 1 | Linear | 500 | ReLU |
| Decoder 2 | Linear | 500 | ReLU |
| Decoder 3 | Linear | 100 | Tanh |

**Table 2.** The hyperparameters used for training the FPM model. The learning rate is the step size for updating the model parameters. The number of epochs is the number of times the model sees the entire training data. The batch size is the number of samples used for each gradient update. The noise factor is the fraction of input values that are randomly corrupted by adding Gaussian noise.

| Hyperparameter | Value |
| --- | --- |
| Learning rate | 0.01 |
| Number of epochs | 100 |
| Batch size | 32 |
| Noise factor | 0.05 |

The loss function is minimized over the reconstructed output and the original input, after adding noise to the input values. The noise helps to prevent overfitting and improve the robustness of the model. The model also incorporates a fairness constraint that penalizes the difference between the distributions of the latent representations for different protected groups, such as gender or race. The fairness constraint helps to reduce the unwanted bias in the model and improve its generalization performance.

**Implementation details:**

We implemented FPM model using Python 3.8.2, with scikit-learn and pytorch as machine learning libraries(34, 35). We used a server with dual 6242R processors and triple RTX 8000 GPUs for training and testing. For brevity, we report only the final setting used in the patient stratification experiments. We performed topic modeling on the clinical notes using LDA, and obtained 100 topics as the optimal number based on perplexity scores. We experimented with different numbers of topics (e.g., 10, 50, 100, 200, 300, 500), but found that 100 topics provided the best balance between interpretability and granularity. We then fed the topic distributions and the structured data to a SDAE, which learned a low-dimensional representation of the patient data. We tuned the hyperparameters of the SDAE to minimize the reconstruction error and ensure fair representation learning. We tested various configurations of the SDAE architecture (e.g., number of layers equal to 1, 3, 5; hidden dimension equal to 100, 200, 300, 400, 500). We found that a three-layer SDAE with 500 hidden units in each layer performed the best in terms of the reconstruction error.